\documentclass[runningheads]{llncs}
\usepackage{graphicx}
\usepackage{tikz}
\usepackage{comment}
\usepackage{amsmath,amssymb} % define this before the line numbering.
\usepackage{color}
\usepackage{amsmath,graphicx}
\usepackage{booktabs}
\usepackage{multirow}
\usepackage{caption}
\usepackage{subcaption}
\usepackage{amsmath}
\usepackage{amssymb}
\usepackage{comment}
\usepackage{float}
\usepackage{booktabs}
\usepackage{cite}
\usepackage[T1]{fontenc}
\usepackage{hyperref}

\begin{document}
\setcounter{secnumdepth}{3}

\pagestyle{headings}
\mainmatter
\def\ECCVSubNumber{100}  % Insert your submission number here

\title{Vision-Based Activity Recognition in Children \\ with Autism-Related Behaviors} % Replace with your title

\titlerunning{Activity Recognition in Children with Autism-Related Behaviours}

\author{Pengbo Wei\inst{1} \and
David Ahmedt-Aristizabal\inst{1,2} \thanks{Same contribution as the first author} %\thanks{Equal contribution by UJ and LW} 
\and
Harshala Gammulle\inst{1,2} \and
Simon Denman\inst{2} \and
Mohammad Ali Armin\inst{1}
}
\authorrunning{Wei et al.}
\institute{Imaging and Computer Vision Group, CSIRO Data61, Canberra, Australia 
\email{\{sam.wei,david.ahmedtaristizabal,ali.armin\}@data61.csiro.au}
\and
SAIVT, Queensland University of Technology, Brisbane, Australia 
\email{\{pranali.gammule,s.denman\}@qut.edu.au}
%\\ \url{http://www.springer.com/gp/computer-science/lncs} 
%\and
%\thanks{joint}
%Software Engineering, CSIRO Data61, Australia\\
%\email{ashley.stacey@data61.csiro.au} 
}
%\end{comment}
%******************
\maketitle

\begin{abstract}
\vspace{-2pt}
Advances in machine learning and contactless sensors have enabled the understanding complex human behaviors in a healthcare setting. 
In particular, several deep learning systems have been introduced to enable comprehensive analysis of neuro-developmental conditions such as Autism Spectrum Disorder (ASD). This condition affects children from their early developmental stages onwards, and diagnosis relies entirely on observing the child’s behavior and detecting behavioral cues. However, the diagnosis process is time-consuming as it requires  long-term behavior observation, and the scarce availability of specialists.
We demonstrate the effect of a region-based computer vision system to help clinicians and parents analyze a child's behavior. For this purpose, we adopt and enhance a dataset for analyzing autism-related actions using videos of children captured in uncontrolled environments (\textit{e.g.}~videos collected with consumer-grade cameras, in varied environments). The data is pre-processed by detecting the target child in the video to reduce the impact of background noise. 
Motivated by the effectiveness of temporal convolutional models, we propose both light-weight and conventional models capable of extracting action features from video frames and classifying autism-related behaviors by analyzing the relationships between frames in a video.
Through extensive evaluations on the feature extraction and learning strategies, we demonstrate that the best performance is achieved with an Inflated 3D Convnet and Multi-Stage Temporal Convolutional Networks, achieving a 0.83 Weighted F1-score for classification of the three autism-related actions, outperforming existing methods. 
We also propose a light-weight solution by employing the ESNet backbone within the same system, achieving competitive results of 0.71 Weighted F1-score, and enabling potential deployment on embedded systems. 
Experimental results demonstrate the ability of our proposed models to recognize autism-related actions from videos captured in an uncontrolled environment, and thus can assist clinicians in analyzing ASD. 

\vspace{-5pt}
\end{abstract}

%----------------------
\section{Introduction}

%**Intro
\label{sec:intro}

% Intro Application domain behaviour analysis in healthcare
A computational understanding of human behaviors has significant potential for applications across multiple domains, including healthcare~\cite{ahmedt2019understanding}. However, modelling and automatically analyzing human behaviors is extremely challenging, as behaviors are contextual and often social, \textit{i.e.} in relation to other people. Thus, an understanding of interaction dynamics may be needed to understand an individual's behavior. An important application area of AI-enabled computational behavior analysis is in characterizing the behavior and developmental changes in children who are diagnosed with autism spectrum disorder (ASD). 

% Intro disorder and justification of the project
ASD is a neurodevelopmental disorder characterized by a set of social communication deficits, self-harm, or persistent repetition of actions \cite{lord2018autism}. Moreover, this disorder often manifests in children during their early developmental stages and can have severe negative impacts on the quality of their life over a long time period \cite{national2001educating}. At present in Australia, in excess of 1 in 150 children are diagnosed with ASD \cite{aihwAutismAustralia}. There can also be a considerable delay in ASD diagnosis from the point at which parents sought professional help \cite{crane2016experiences}. 
There are two reasons for the long wait times for ASD diagnosis; i) the low availability of specialists, ii) there are no reliable biomarkers for ASD, and its diagnosis requires a long-term observation of stereotypical behaviors such as headbanging, arm-flapping or spinning. 
It is worth noting that in this study, we only analyze stereotypical behaviors and other ASD related biomarkers such as those determined through fMRI, facial expressions, eye gaze, and motor control/movement patterns are outside the scope of this study. Interested readers are referred to ~\cite{de2020computer} for more information.

%Intro traditional domain application in ASD
Computational methods have been effectively employed to assist with the diagnosis of ASD, and automatically assess ASD behavioral biomarkers through video analysis (\textit{e.g.}~temporal ~\cite{zunino2018video,kojovic2021using,ali2022video} or static analysis~\cite{liang2021autism}). 

%Preliminary work dynamic (children data)~\cite{rajagopalan2014detecting}
A pioneering computer vision based method for analyzing self-stimulatory behaviors for ASD from video was proposed by Rajagopalan et. al~\cite{rajagopalan2013self}. They developed a system based on Bag of Words (BOW). Subsequently, in ~\cite{rajagopalan2014detecting}, they modelled self-stimulatory behaviors by selecting poselet bounding boxes and a high-level global descriptor that captured dominant motion patterns in video.
Recently in \cite{kojovic2021using}, the authors used spatial-temporal model using VGG16 \cite{simonyan2014very} and LSTMs to analyze ASD behaviors by analyzing the 2D pose extracted from sequential frames. However, in a typical video dataset, some samples often include additional irrelevant people. The presence of these people can hamper the action recognition model during training. Thus, in \cite{washington2021activity}, they extract the target head pose from videos with a head banging detector, and then analyze the head banging action using a CNN+LSTM architecture.
Similarly, in \cite{ali2022video}, 3D Convolutional Neural Networks (3D-CNN) along with target person detection and tracking techniques are adopted for action recognition tasks in videos of children with ASD. In \cite{negin2021vision}, they investigated two approaches and compared a bag-of-visual-words approach with RNNs and CNNs, and showed that the deep learning architectures offer competitive performance for the classification of four actions including arm flapping, headbanging, hand actions and spinning. Furthermore, in \cite{pandey2020guided}, to solve issues related to data shortages the authors proposed a guided weak supervision method to match the target data to a class in the source data, and subsequently trained a classifier on the augmented data to recognize autism-related behaviors. 

%Static Analysis
While the dynamic analysis of videos (where the temporal nature of the data is expressly modelled) is more common, some researchers have developed static approaches for ASD behavioral analysis. In \cite{liang2021autism}, the authors used an unsupervised model they called Temporal Coherency Deep Networks (TCND) that consists of four ALexNet~\cite{krizhevsky2012imagenet} models with the same parameters to extract features from frames, after which a linear SVM is used to classify actions based on the extracted features. This is an example of a static approach which disregards temporal features and classifies an individual video frame. Results at the video level are reported based on the average obtained by applying the framework to each frame.     

In general, Temporal Convolutional Networks (TCN) have shown better performance in comparison to LSTMs for various tasks including action recognition \cite{bai2018empirical}. In \cite{wang2019i3d}, the authors showed that the combination of a 3D-CNN and temporal models can increase Human Action Recognition performance. However, the researchers focused on adopting very highly parameterized models to analyze ASD behaviors, thus limiting the ability to deploy such a method on an embedded device. To overcome this limitation, the method introduced in this paper focuses on light-weight models for the classification of ASD behaviors in complex environments, such as from videos collected with consumer grade cameras.

% introducing dataset and challenges
Apart from aforementioned framework challenges, an appropriate dataset is another important component needed to address ASD monitoring. Rajagopalan et al.~\cite{rajagopalan2013self} introduced a new Self-Stimulatory Behavior Dataset (SSBD) for ASD. This dataset was collected from YouTube videos recorded in uncontrolled environments, and included three stereotypical behaviors including arm flapping, headbanging and spinning. Compared with the traditional action recognition datasets which consists of tens of thousands of action videos(e.g. Kinetics~\cite{kay2017kinetics}), our dataset that contains 61 videos can be considered a shallow dataset. 
Considering the small size of this dataset, Negin et al.~\cite{negin2021vision} introduced another dataset with an extra behavior (hand action) collected from YouTube, and named this the Expanded Stereotype Behavior Dataset (ESBD). This dataset consists of 141 video and has no shared videos with the SSBD dataset. However, this dataset has not yet been made publicly available. 

%Our proposal
Considering the lack of a dataset that captures realistic video from subjects in an unconstrained environment, we introduce a stereotypical behavior dataset which is an extension of SSBD. We add newly collected autism-related videos from online media platforms to the existing SSBD. Utilizing this extended dataset, we develop a spatio-temporal framework for recognizing autistic behaviors from videos. Our proposed method consists of two main streams: a feature extractor that leverages a temporal 3D CNN to extract embedding features from autism video frames, and a temporal model that detects and classifies each action based on the extracted embeddings.   

Our main contributions are summarized as follows:
\begin{enumerate}
\vspace{-2pt}
\item We extend an existing dataset of stereotypical children’s autism behavior videos recorded in an uncontrolled environment by incorporating 12 new subjects, and removing 11 original noisy subjects. The new dataset contains 61 videos and is divided into 168 short video clips. Information of each video (URL and type) is available at 
\href{https://github.com/Samwei1/autism-related-behavior/blob/main/url_list.pdf}{https://github.com/Samwei1/autism-related-behavior}.
\item We exploit region-based action recognition frameworks that utilize conventional and light-weight architectures combined with multi-stage temporal convolutional networks to recognize characteristic autism behaviors, achieving superior performance in comparison to baseline models. A compressed feature representation is evaluated to ascertain if real-time throughput with acceptable accuracy is possible.
\end{enumerate}
%\vspace{-3pt}

%*************************************************************************************************
\section{Methods} 

\label{methodology}

\begin{figure*}[!t]
	\centering
    \includegraphics[width=0.9\linewidth]{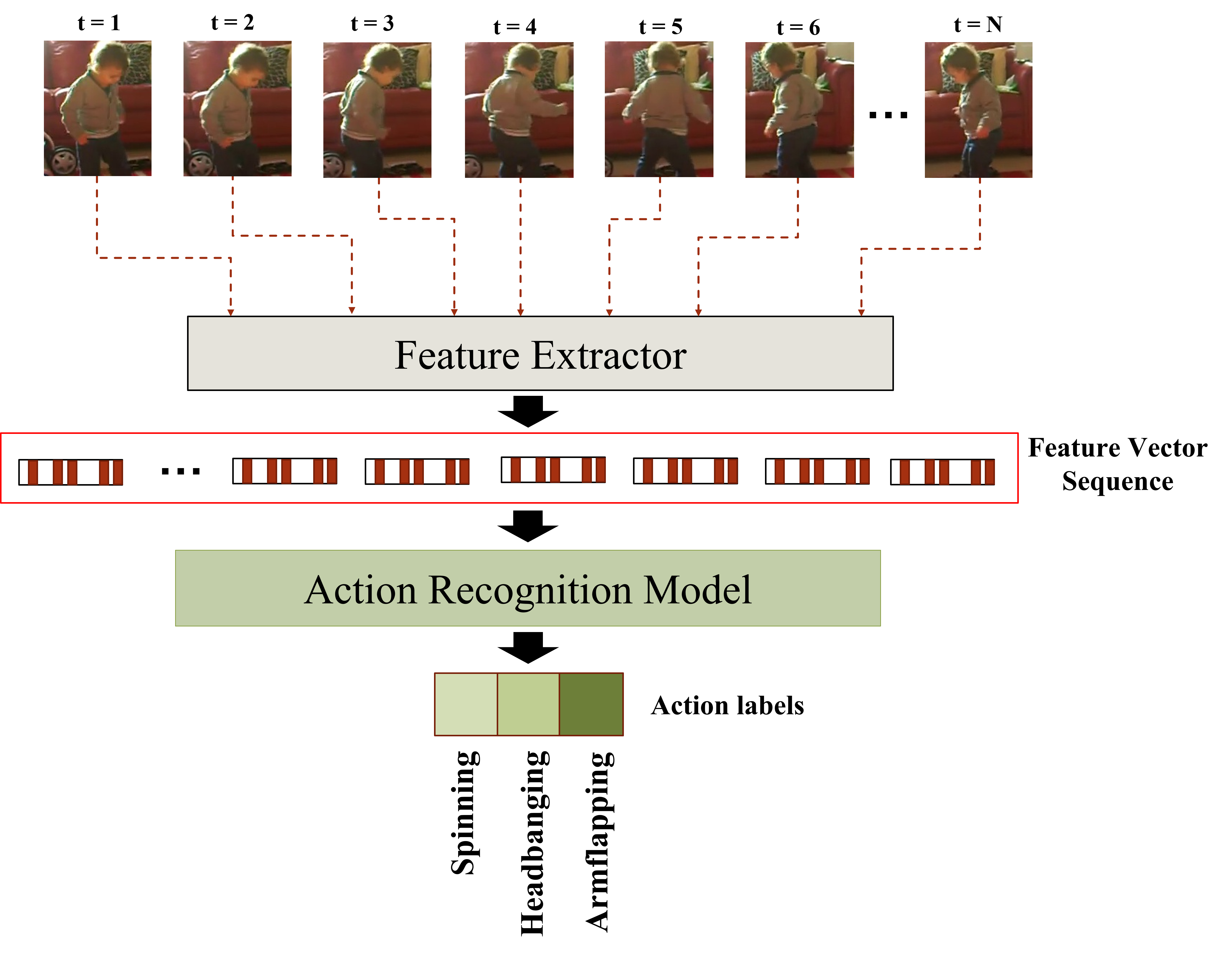}
    \vspace{-8pt}
	\caption{Overview of the activity recognition pipeline. Given a sequence of RGB video frames, the system generates a sequence of feature vectors via the feature extractor. Next, the action recognition model recognizes the activity by using the generated sequence of visual features. }
	\label{fig1}
\end{figure*}

In this section, we describe the proposed region-based deep learning approach that recognizes autism-related behaviors in children. Figure~\ref{fig1} illustrates the overall architecture of our framework, which has two crucial steps:  
\textbf{i}) a feature extractor that extracts frame-wise action-related semantic features; and 
\textbf{ii}) an action recognition model which recognizes autism-related behaviors by modelling the temporal evolution of the extracted frame-wise features.
We empirically compare the performance of four deep learning architectures with respect to the utilized feature extraction models.
In the following subsections, we provide a more detailed explanation of the feature extraction and the action recognition methods that were used in our experiments. We further investigate different combinations of these feature extractors and action recognition models to recognizing autism-related behaviors efficiently.

%-----------
%-----------
\subsection{Feature extraction backbones}
\label{subsec:fe}

Feature extraction is a crucial step in the proposed action recognition pipeline which enables the extraction of informative features from the raw data. It also helps in reducing the amount of redundant data fed to the action recognition component, which further enhances the action learning process.

Traditional feature extraction methods (such as spatio-temporal interest points (STIP) used in \cite{rajagopalan2013self}) require feature engineering and rely on substantial domain knowledge of human experts to obtain best results. In contrast, deep learning models learn features automatically and have been used to achieve better performance for feature extraction and classification tasks \cite{tan2019efficientnet, xie2017aggregated, howard2017mobilenets, gu2018recent}.
Once the inputs and the corresponding labels are given, deep networks are able to learn a mapping that captures the relationship between the input and the output data automatically. However, learning this mapping requires a great deal of data and time to train a deep network from scratch. When the requisite amount of data is not available, a common practice is to adopt transfer learning, where we reuse a model that has previously been trained on a large dataset and fine-tune the network (or a portion of the network) with our target dataset. 
Considering these approaches, in our work, we investigate using both pre-trained and fine-tuned feature extractors to determine the best feature extractor for the behavior recognition task.

We select feature extractors appropriate for two different operating conditions: 
i) \textit{a light-weight environment} condition suitable for an embedded device; and ii) \textit{a standard operating} condition (\textit{i.e.} conventional models) suitable for a powerful workstation or server. 
We investigated the following feature extractors:
EfficientNets~\cite{tan2019efficientnet}
MobileNet~\cite{howard2017mobilenets},
ShuffleNet~\cite{ma2018shufflenet},
Enhanced ShuffleNet (ESNet)~\cite{yu2021pp}, 
ResNet~\cite{he2019bag}, 
%DenseNet121 (8M)
and Inflated 3D ConvNet (I3D)~\cite{carreira2017quo}. 
The models EfficientNet-B0, MobileNetV3, ShuffleNetV2 and ESNet are identified as light-weight models (number of model parameters $<$ 5.5M), while EfficientNet-B3, ResNet18 and I3D are considered as the standard operating condition (number of model parameters $<$ 15M). 
It should be noted that the \textit{standard operating condition} models themselves are still substantially less computationally demanding compared to the two-stream models utilized in \cite{ali2022video}, or more complex networks in the literature such as ResNet101\_vd (44M), HRNet\_W48 (78M), and SwinTransformer (99M).
In \cite{ali2022video}, the authors use I3D models for both RGB and Optical flow inputs, whereas in our work, only a single I3D model is applied to the RGB frames. 

When choosing feature extraction models, we focused on selecting light-weight models with a low number of trainable parameters. The main reason for this is to support a real-world application by improving the efficiency of the feature extraction step and helping to reduce data requirements for fine-tuning. In addition, deeper networks can struggle during training due to problems such as vanishing gradients caused by the increased network depth. The following subsections discuss the technical details of the feature extraction networks (backbones).

%---------------
\subsubsection{EfficientNets} 
EfficientNets\cite{tan2019efficientnet} are composed of a deep CNN architecture with fewer parameters compared to previous models such as VGG and AlexNet. EfficientNets are scaled by an effective compound scaling mechanism, where the network is uniformly scaled in all dimensions including depth, width, and resolution, from a baseline EfficientNet architecture that is generated by a neural architecture search. 
Among their models, EfficientNet-B0 is the simplest and most efficient model. Despite being light-weight, EfficientNet-B0 has achieved the state-of-the-art performance on image classification tasks with fewer parameters on average than other CNN models, such as AlexNet\cite{krizhevsky2012imagenet}, Inception-v2\cite{szegedy2016rethinking}, and ResNet-50\cite{he2016deep}. 
We also consider EfficientNet-B3, which achieves improved performance over previous architectures but with fewer parameters (81.1\% accuracy on ImageNet\cite{deng2009imagenet} with 12M parameters). Critically, the transfer learning performance of EfficientNet has also been shown to be better than other architectures \cite{tan2019efficientnet}.

%---------------
\subsubsection{MobileNets} 
MobileNets are light-weight CNN models based on a streamlined architecture that uses depth-wise separable convolutions to build light weight yet deep neural networks. They are widely used in mobile and embedded vision applications and achieve relatively high performance for object detection, fine-grained classification, face attribute classification, and large scale geo-localization tasks \cite{howard2017mobilenets}.
In this research, we adopt MobileNetV3~\cite{howard2019searching} which is more accurate for image classification tasks than previous versions of MobileNet, while also being more computationally efficient. 

%---------------
\subsubsection{ShuffleNet} 
ShuffleNet~\cite{zhang2018shufflenet} is a light-weight model that is widely adopted in portable and low computation applications. To solve issues that exist in other light-weight networks (a larger number of feature channels causing high computation complexity), Zhang et al.\cite{zhang2018shufflenet} proposed two techniques, pointwise group convolutions and bottleneck-like structures. However, the two techniques were found to be costly for light-weight networks. To further reduce the computation complexity of ShuffleNet, Ma et al.\cite{ma2018shufflenet} introduced the channel split operator to help maintain a large number and equally wide set of channels while avoiding dense convolution and having too many groups of convolution operations in networks, and thus proposed the ShuffleNetV2.

%---------------
\subsubsection{ESNet} 
The Enhanced ShuffleNet (ESNet)~\cite{yu2021pp} further improves the network structure of ShuffleNetV2~\cite{ma2018shufflenet} for real-time object detectors, with a better balance between accuracy and latency on mobile devices. 
ESNet adapts some methods used by the lightweight CPU network PP-LCNet~\cite{cui2021pp}, with additional modifications.
The Squeeze-and-excitation module (SE)~\cite{hu2018squeeze} is a channel attention module, which is also used in MobileNetv3, and is included in all blocks of ESNet to improve the feature extraction by weighting the network channels. Furthermore, the author of GhostNet~\cite{han2020ghostnet} proposes a novel Ghost module that can generate more feature maps with fewer parameters to improve the network’s learning ability. The Ghost module is also added in the blocks with stride set to 1 to further enhance the performance of the ESNet.

%---------------
\subsubsection{ResNet} 
Recent studies have shown the high performance of ResNet~\cite{he2016deep}, which leverages the residual block structure, for image recognition and classification tasks. It uses residual (or skip) connections to allow information to more readily propagate through the network.
To improve the accuracy and robustness of this popular network without increasing the amount of computation, the authors in~\cite{he2019bag} proposed an improved method
ResNet-vd with a collection of refinements, such as modifying the downsampling operation, which can generate better feature maps. ResNet models have many variants with different number of layers and different residual block structures. In this study we investigated ResNEt18-vd which consists of 18 residual blocks and 11.7M parameters. 
%The downsampling operation through convolution 1 $\times$ 1 with stride = 2 is transferred to the convolution of 3 $\times$ 3, and the downsampling operation through convolution 1 $\times$ 1 is canceled, and the average pooling layer is added to downsampling the feature map. 

%---------------
\subsubsection{Inflated 3D Convnet (I3D)} 
Inception-VI was used as the backbone network for the I3D model \cite{carreira2017quo} by expanding the 2D convolution filters and pooling kernels to 3D, allowing the model to learn spatial-temporal features from videos. Thus, the typical square filters become cubic, with an N $\times$ N filter becoming one of size N $\times$ N $\times$ N. 
The pre-trained I3D model trained on the Kinetics dataset \cite{carreira2017quo} achieved the state-of-the-art performance for action classification. In many previous works \cite{ali2022video,farha2019ms}, feature extraction is performed using two streams of data (RGB and optical flow), with each stream processed by an independent I3D model. In our work, we only used the RGB stream of I3D instead of both streams to obtain a less complex feature extractor.

%-----------
%-----------
\subsection{Action recognition models}
\label{subsec:ar}
Autism related behaviors are dynamic events, and last for a variable duration of time. Hence, it is necessary to capture temporal information to model these behaviors and accurately classify them.
The deep features extracted using the approaches outlined in Section \ref{subsec:fe} are used as inputs to a temporal model which is trained to classify actions. We select the following four prominent temporal deep learning approaches, the LSTM and three temporal convolutional network methods.

%---------------
\subsubsection{Long short-term memory (LSTM)} 
The LSTM~\cite{hochreiter1997long} is a type of recurrent neural network (RNNs) that is capable of learning temporal relationships from time series data. 
Compared with traditional RNN models, LSTMs address the vanishing gradient problem when the temporal dimension becomes large. There are three logic gates (an input gate, an output gate, and a forget gate) in a LSTM cell that control the flow of information into and out of the cell. The input gate controls which information should be stored in the hidden memory, while the forget gate controls which information in the memory should be forgotten from the previous cell state. The output gate is responsible for determining which information should be output of the LSTM at a given timestep. In this work, we use an LSTM network with the configuration shown in Table \ref{table:lstm}.

\begin{table}[!t]
\caption{Configuration of LSTM architecture}
\centering
\resizebox{0.7\textwidth}{!}{%
\label{table:lstm}
\begin{tabular}{c c} 
 \hline
\textbf{Layers}  & \textbf{Configuration} \\
 \hline
 \hline
Layer1 &  bidirectional LSTM layer with 512 hidden neurons \\
 \hline
Layer2 &  bidirectional LSTM layer with 512 hidden neurons \\
 \hline
Layer3 &  bidirectional LSTM layer with 512 hidden neurons \\
 \hline
Layer4 &  Dense(128), with Relu activation function\\
 \hline
Layer5 &  Dense(3), followed by softmax function\\
 \hline
\end{tabular}}
\end{table}

\begin{table}[!t]
\caption{Configuration of TCN architecture}
\centering
\resizebox{0.7\textwidth}{!}{%
\label{table:tcn}
\begin{tabular}{c c} 
 \hline
\textbf{Layers}  & \textbf{Configuration} \\
 \hline
 \hline
Layer1  &  dilation TCN layer with kernel size = 5, level size = 5  \\
 \hline
Layer2 &  Dense(256), with Relu activation function\\ 
 \hline
Layer3 &  Dense(3), followed by log softmax function\\ 
 \hline
\end{tabular}}
\end{table}

%---------------
\subsubsection{Temporal Convolutional Networks (TCN)} 
The TCN \cite{bai2018empirical} is a temporal model proposed for video-based action segmentation and detection tasks. Due to their capacity to learn temporal relations among feature sequences, TCNs have been widely used for various video-based problem domains that seek to obtain a single \cite{gammulle2021multi} or sequence of prediction outputs \cite{chen2020action,gammulle2021tmmf}. In this work, we adapt the TCN architecture introduced in \cite{bai2018empirical} to recognize autism-related activities performed by children.
TCNs are different to RNN models based on the following characteristics: 
1) Causal convolution, meaning that the value returned by the model at the current step only depends on information from previous observations, and there is no information flow from the future to the past. 
2) Dilated convolutions, which enables TCN models to achieve an exponentially large receptive field. That means the TCN model can have a large receptive field, even with few layers. 
3) Residual connections, which have been shown to be an effective method to address vanishing gradient issues that exist in deep neural networks, and allow the network to pass information in a cross-layer manner. The details of the TCN configuration used in this work is shown in Table \ref{table:tcn}.

%---------------
\subsubsection{Multi-Stage Temporal Convolutional Network (MS-TCN)}

\begin{figure*}[!h]
	\centering
    \includegraphics[width=0.7\linewidth]{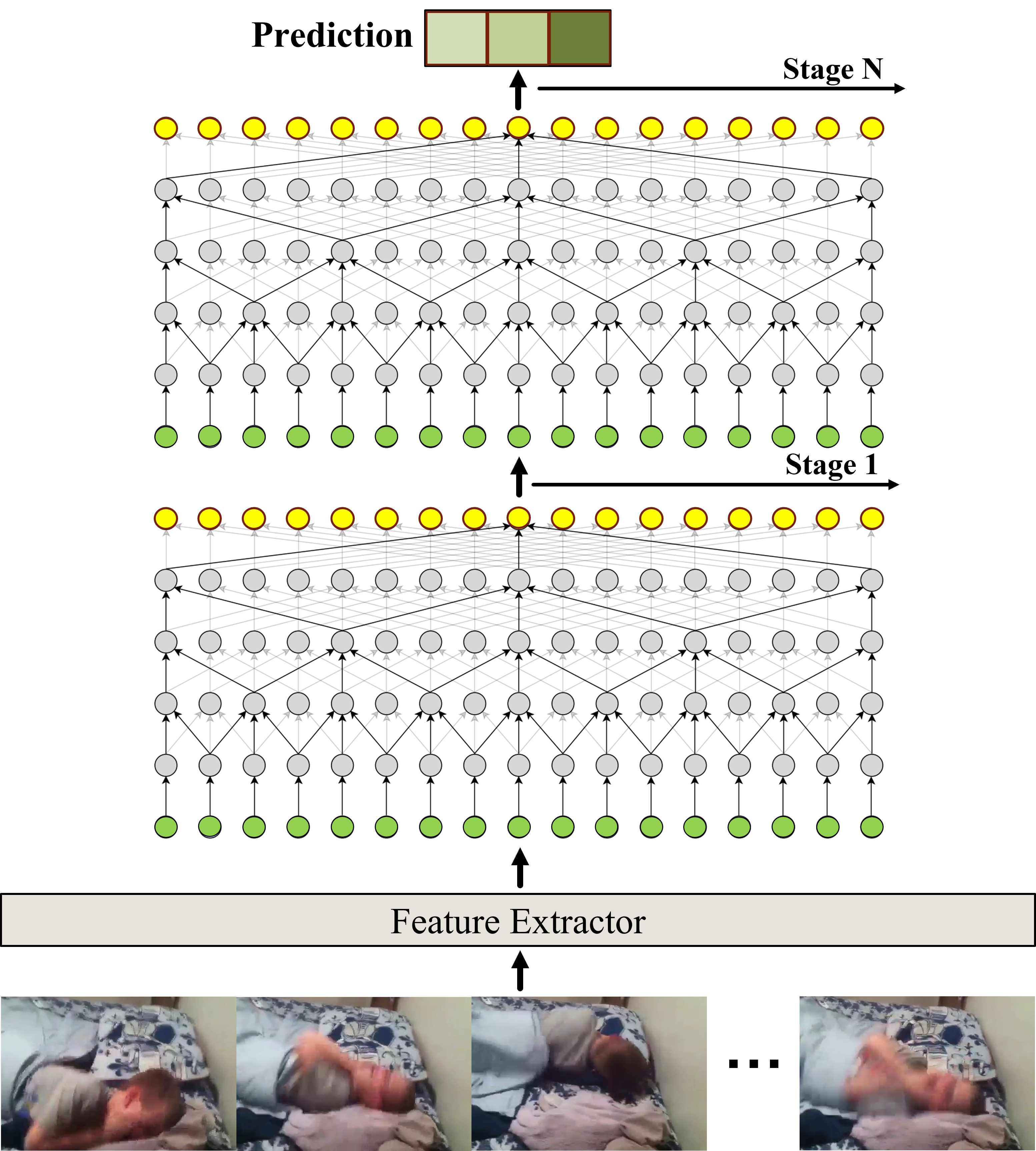}
    \vspace{-8pt}
	\caption{Multi-stage temporal convolutional network. Each stage produces an initial prediction, which is refined by the subsequent stage. Recreated from~\cite{farha2019ms}.}
	\label{fig2}
\end{figure*}

Motivated by the success of the TCN architecture introduced in ~\cite{lea2017temporal}, in~\cite{farha2019ms} the authors proposed a multi-stage temporal convolution network the temporal action segmentation. The MS-TCN architecture is illustrated in Figure~\ref{fig2}. At each stage of MS-TCN, an initial prediction is generated, which is then refined by the next stage. Each stage is a single-stage TCN model which is composed of a $1 \times 1$ convolution layer followed by several dilated 1-dimensional convolutional layers. Inspired by \cite{oord2016wavenet}, the dilation layers are designed with a dilation factor that is doubled at each subsequent layer (\textit{i.e.} 1, 2, 4,$\dots$, 512). This model does not use any pooling or fully connected layers in order to maintain the temporal resolution, and thus results in a lower trainable parameter count than other methods. We adopt the MS-TCN architecture for the behavior recognition task as shown in Figure \ref{fig2} and the details of the TCN configuration used in this work is shown in Table \ref{table:ms_tcn}.

\begin{table}[!t]
\caption{Configuration of MS-TCN architecture}
\centering
\resizebox{0.7\textwidth}{!}{%
\label{table:ms_tcn}
\begin{tabular}{c c} 
 \hline
\textbf{Layers}  & \textbf{Configuration} \\
 \hline
 \hline
Layer1 &  dilation TCN layer with kernel size = 5, level size = 5  \\
 \hline
Layer2 &  dilation TCN layer with kernel size = 5, level size = 5  \\
 \hline
Layer3 &  dilation TCN layer with kernel size = 5, level size = 5  \\
 \hline
Layer4 &  dilation TCN layer with kernel size = 5, level size = 5 \\
 \hline
Layer5 &  dilation TCN layer with kernel size = 5, level size = 5 \\
 \hline
Layer6 &  Dense(256), with Relu activation function\\ 
 \hline
Layer7 &  Dense(3), followed by log softmax function\\ 
 \hline
\end{tabular}}
\end{table}

\subsubsection{Extended Multi-Stage Temporal Convolutional Network (MS-TCN++)}
\begin{figure*}[!t]
	\centering
    \includegraphics[width=0.8\linewidth]{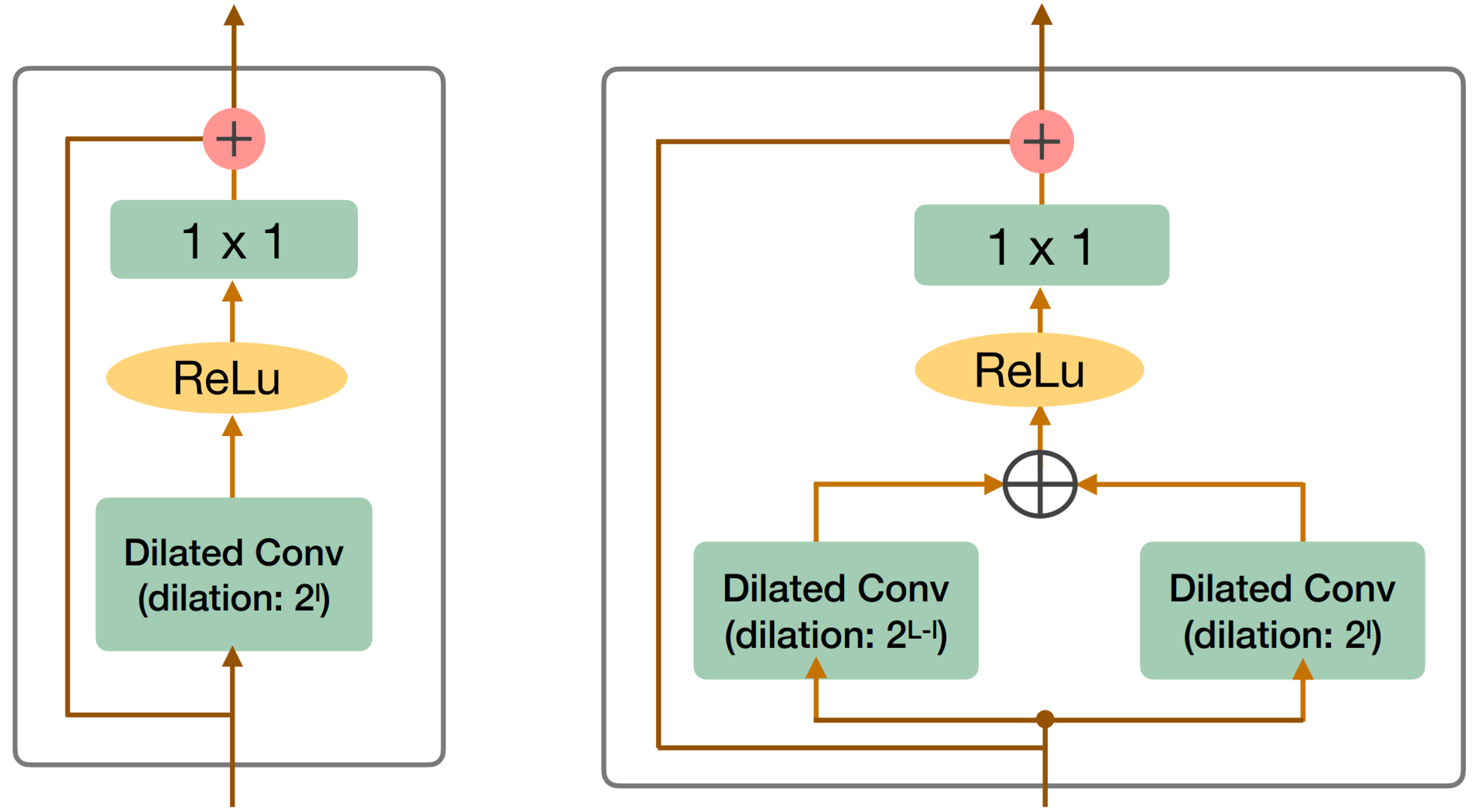}
    \vspace{-8pt}
	\caption{A comparison between MS-TCN and MS-TCN++. A presentation of the dilated residual layer in MS-TCN (Left) and a dual dilation residual layer in MS-TCN++ (Right). Adapted from~\cite{gammulle2022continuous}.}
	\label{fig:MS_TCN_plus}
\end{figure*}

MS-TCN++ is an extension of the MS-TCN. Compared to the MS-TCN, the MS-TCN++ has a dual dilation layer that overcomes issues related to the lower layers having too small a receptive field. The dual dilation layer is depicted in Figure~\ref{fig:MS_TCN_plus}. Similar to the MSTCN, the MS-TCN++ architecture is adopted here for the recognition of autism related behavior. The details of the TCN configuration used in this work is shown in Table \ref{table:ms_tcn_extend}.

%---------------
\begin{table}[!t]
\caption{Configuration of MS-TCN++ architecture}
\centering
\resizebox{0.7\textwidth}{!}{%
\label{table:ms_tcn_extend}
\begin{tabular}{c c} 
 \hline
\textbf{Layers}  & \textbf{Configuration} \\
 \hline
 \hline
Layer1 &  dual dilation TCN layer with kernel size = 5, level size = 5  \\
 \hline
Layer2 &  dual dilation TCN layer with kernel size = 5, level size = 5  \\
 \hline
Layer3 &  dual dilation TCN layer with kernel size = 5, level size = 5  \\
 \hline
Layer4 &  dual dilation TCN layer with kernel size = 5, level size = 5 \\
 \hline
Layer5 &  dual dilation TCN layer with kernel size = 5, level size = 5 \\
 \hline
Layer6 &  Dense(256), with Relu activation function\\ 
 \hline
Layer7 &  Dense(3), followed by log softmax function\\ 
 \hline
\end{tabular}}
\end{table}

%*************************************************************************************************
\section{Experimental setup} 
\label{experiments}

\begin{figure*}[t!]
	\centering
    \includegraphics[width=0.9\linewidth]{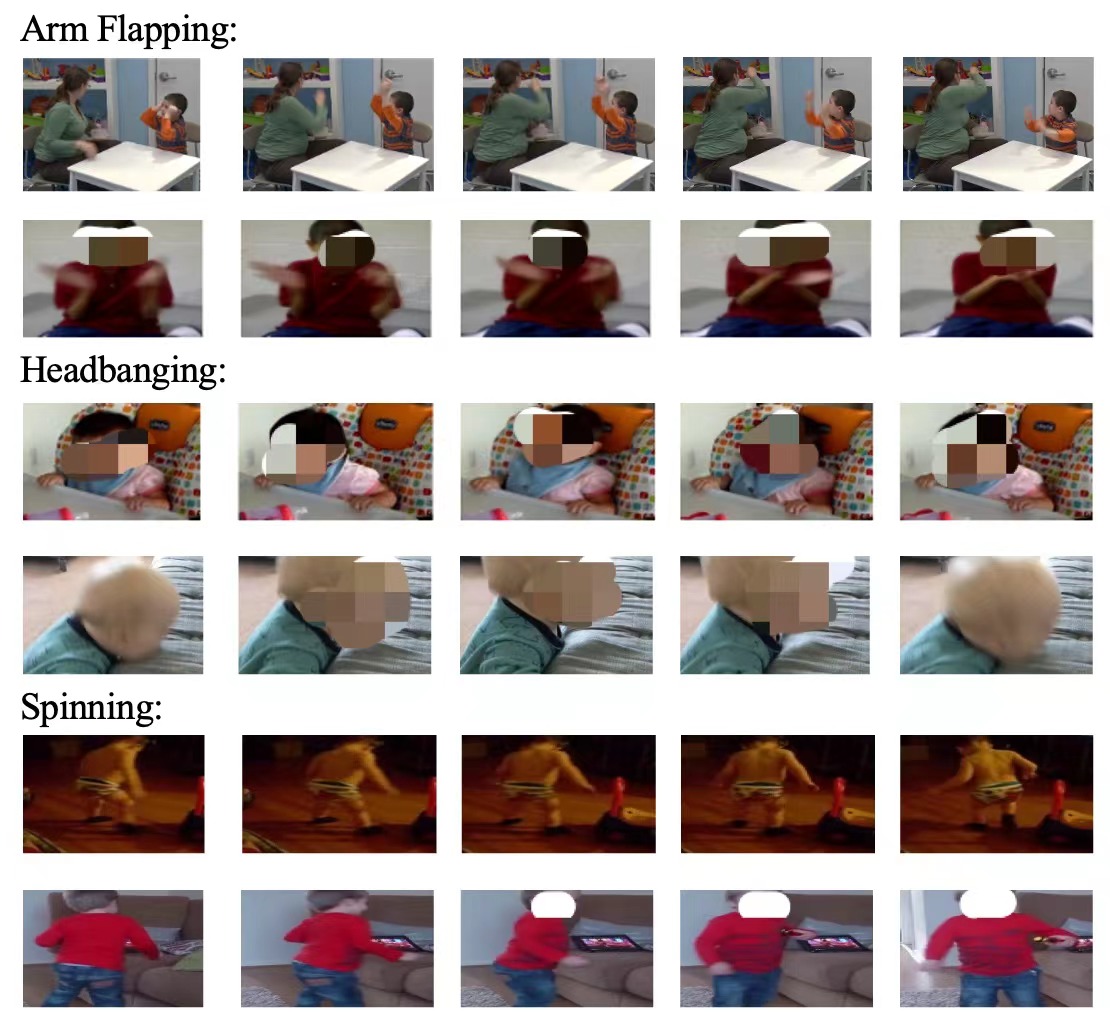}%
	\caption{Selected frames from  the videos in the dataset used. From top to bottom, the autism-related behaviors 'arm flapping', 'headbanging', and 'Spinning' are shown. These videos were recorded in an uncontrolled environment, and some videos contain activities that involve human interactions (\textit{e.g.}~the sample in the first row).}
	\label{fig3}
\end{figure*}

%-----------
\subsection{Dataset and preprocessing}
\label{sec:trainingdatasets} 

Due to ethical issues, publicly available datasets showing autism behavior are limited (we note that the dataset proposed by Negin et al.~\cite{negin2021vision} is not yet publicly available). We train and test all models on a modified version of the publicly available self-stimulatory behavior dataset (SSBD)~\cite{rajagopalan2013self}.  
The videos in this dataset are diverse in nature and are recorded in uncontrolled environments, and collected from different online portals such as YouTube, Vimeo, and Dailymotion. Selected samples are depicted in Figure~\ref{fig3}.
There are three stereotypical autism behaviors within the data: arm flapping, headbanging, and spinning. 
The complete set of videos reported by~\cite{rajagopalan2013self} comprises 75 videos, of which only 60 are downloadable due to privacy concerns.

\begin{figure*}[htbp]
	\centering
    \includegraphics[width=0.95\linewidth]{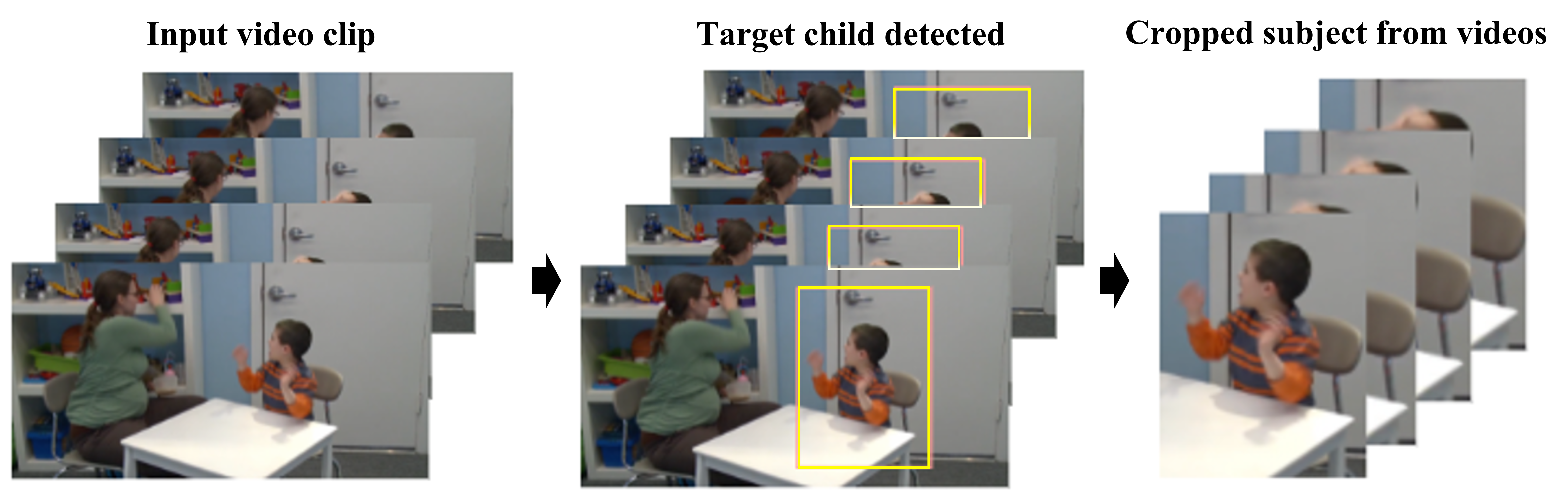}%
	\caption{The pre-processing procedure: The child of interest is detected using the Detectron2 \cite{wu2019detectron2} human detector, after which the target child is cropped from the video. }
	\label{fig4}
\end{figure*}

% Data curation:
In this dataset, some videos are very noisy and captured in very dark environments. Thus, we collected new videos from YouTube to augment the original dataset following the methodology proposed by the authors of the SSBD dataset. We manually curated the videos and categorized them into their corresponding autism-related behaviors. The final version of the dataset used on our experiments is composed of 61 videos and 61 subjects, and consists of 20, 21 and 20 videos for arm flapping, headbanging, and spinning, respectively. A comparison between the original and updated SSBD dataset is presented in Table~\ref{table:0}

%Original dataset -- video clips, subjects per class
%Curated dataset
\begin{table}[!t]
\caption{Comparison between the original and updated SSBD dataset}
\centering
\resizebox{0.9\textwidth}{!}{%
\label{table:0}
\begin{tabular}{c c c c} 
\toprule
& Armflapping & Headbanging & Spinning \\
\midrule
No. Videos in original SSBD dataset~\cite{rajagopalan2013self} & 23  &19 & 18 \\
No. Videos in the updated SSBD dataset & 20  & 21 & 20 \\
\bottomrule
\multicolumn{4}{p{350pt}}
{
Information of the source of data is available at
\href{https://github.com/Samwei1/autism-related-behavior}{Github Samwei1}  \newline
} %\newline
\end{tabular}}
\vspace{-10pt}
\end{table}

%Preprocessing
For data preparation, we segmented videos into short clips to increase the number of samples. We fix the number of extracted key-frames from each sample video to 20 frames, which is sufficient to cover the complete length of each action. Raw frames from all video clips are extracted with FFmpeg\footnote{https://ffmpeg.org/.}. After performing the data preparation, the total number of samples for each behavior in the ASD dataset were 57, 53, and 58 for arm flapping, headbanging and spinning, respectively.

To reduce the impact of the noisy environment captured within the data, we pre-processed the videos before training by detecting and cropping the target child from videos. In this way, irrelevant parts of videos are ignored, helping the proposed model analyze behaviors by learning more relevant visual features. 
For this purpose, we adopted the Mask-RCNN framework\cite{he2017mask}, which is a state-of-the-art object detector, to localize the target child and obtain a corresponding detection confidence score.
We used the pre-trained weights obtained by training on the COCO dataset \cite{lin2014microsoft}, and the implementation in Detectron2 \cite{wu2019detectron2}. 
Finally, we crop the raw images based on the detected bounding boxes, resize all frames to a fixed resolution based on the input size to the feature extractor, and pass these frames through the feature extractor.
A representation of this pre-processing phase is depicted in Figure~\ref{fig4}.

We developed an experimental dataset to augment the feature extraction process, and demonstrate the utility of fine-tuning feature extractors on a sub-set of training data, rather than merely using pre-trained feature extractors (\textit{e.g.} pre-trained on ImageNet\cite{deng2009imagenet}). The Kinetics dataset~\cite{kay2017kinetics} is a large action recognition dataset and contains some actions that are similar to the actions in our dataset. We combined action videos from Kinetics dataset with the training set videos from the updated SSBD dataset to construct a new dataset to help the feature extractors better understand the ASD behaviors. We named this dataset the \textit{customized Kinetics dataset}.

%-----------
\vspace{-4pt}
\subsection{Evaluation protocol and implementation}
To evaluate our proposed spatio-temporal approaches, we use a 5-fold cross-validation, and report average results across the five folds. To reduce the effects of subject variance, we split the dataset in a subject-wise manner, meaning that any subject will exist only a single fold of the dataset, and thus the subjects used for training and testing for a given fold are disjoint.

As this is a multi-class classification task, weighted F1-score is used as it considers each class's support. This metric can be calculated as follows: 

\[ Weighted\:F1 score = \sum_{i=1}^{3} \frac{2 \times precision_i \times recall_i }{precision_i + recall_i} \times w_i \]

where $w_i$ is the weight of the i-th class, i.e. the percentage of the total samples that belong to the i-th class.

For action recognition models, we trained all proposed models from scratch by using the extracted spatial features in each frame from the feature extractor. 
Models were trained for 200 epochs with a batch size of 16. Categorical cross-entropy loss and the Adam optimizer~\cite{kingma2014adam} (with a learning rate of $1e-3$) were used during training. These hyper-parameters are fine-tuned based on the performance of the models on the validation set of 5-fold cross-validation.

%*************************************************************************************************
\section{Experimental results and discussion}
We evaluated the performance of different combinations of features extractors and action recognition models for the detection of autism related behaviors. 

%-----------
\subsection{Feature extractor evaluation}
\label{sec:backbone_best}

We investigated the performance of different feature extractors (light-weight and conventional) by comparing the performance of all feature extractors using a common TCN as the action recognition model. 
The light-weight feature extractors include MobileNet-V3, ShuffleNetV2, EfficientNet-B0, and ESNet. These are of more relevance given the final goal of adapting the models to a real-world application. The conventional feature extractors are ResNet18\_vd, EfficientNet-B3 and the RGB stream I3D. %The first step is to select a backbone for each operating condition that offers better understanding of the action features.

For the particular case of light-weight backbone models, we compare the performance of the models to extract relevant action features from video frames using the pre-trained feature extractors trained on ImageNet, and those fine-tuned with our customized Kinetics dataset as outlined in Section \ref{sec:trainingdatasets}. Experimental results on two selected backbones are shown in Table~\ref{table:3}. As shown in Table~\ref{table:3}, the fine-tuning of backbones leads to a slight improvement in accuracy of the action recognition model. Considering this improvement, we carried out all experiment of the light-weight backbones by first fine-tuning the feature extraction model.

All conventional feature extractors adopted were only pre-trained on the original Kinetics dataset, as we observed fine-tuning instability when using the modified Kinetics dataset due to the small size of the SSDB dataset and the complexity of the backbone, which resulted in catastrophic forgetting in some cases. The performance and the size of each feature extractor analyzed is shown in Table~\ref{table:1}.
The results in Table~\ref{table:1}, show that light-weight feature extractors have lower performance in comparison to the conventional and more complex feature extractors. Among light-weight models, ESNet achieved the best performance with a weighted F1-score of 0.63, while for the complex models the RGB I3D backbone achieved the best weighted F1-score 0.75. 

%-----
\begin{table}[!t]
\caption{Five-fold evaluation results on the modified SSBD dataset using the fine-tuned feature extractor}
\centering
\resizebox{0.65\textwidth}{!}{%
\label{table:3}
\begin{tabular}{lc}
\toprule
\textbf{Models}  &  \textbf{Weighted F1-score}   \\ 
\midrule
ShuffleNetV2$\ast$ + TCN  & 0.54  \\
ShuffleNetV2$\diamond$ + TCN  & 0.55  \\
ESNet$\ast$ + TCN  & 0.61  \\
ESNet$\diamond$ + TCN  & 0.63  \\
\bottomrule
\multicolumn{2}{p{230pt}}
{ 
$\ast$ Pre-trained on ImageNet. \newline
$\diamond$ Fine-tuned on customized Kinetic dataset.
}
\end{tabular}}
%\vspace{-6pt}
\end{table} 
%-----

%-----
\begin{table}[!t]
\caption{Five-fold evaluation results on the modified SSBD dataset, using all feature extraction models. The number of parameters for each model is also reported.}
\centering
\resizebox{0.8\textwidth}{!}{%
\label{table:1}
\begin{tabular}{lcc}
\toprule
Backbone  &  Weighted F1-score & Parameters   \\ 
\midrule
MobileNet-V3$\star$ + TCN  & 0.50 & 5.4M \\
ShuffleNetV2$\star$ + TCN & 0.55 & 1.17M \\
EfficientNet-B0$\star$ + TCN  & 0.59  &  5.3M \\
ESNet$\star$ + TCN & \textbf{0.63} & 2.1M \\
\midrule
ResNet18\_vd$\dagger$ + TCN  & 0.52 & 11.72M  \\
EfficientNet-B3$\dagger$ + TCN  & 0.67  & 12M  \\
RGB I3D$\dagger$ + TCN  & \textbf{0.75} & 12.7M  \\
\bottomrule
\multicolumn{2}{p{230pt}}
{ 
$\star$ Pre-trained on customized Kinetics dataset. \newline
$\dagger$ Pre-trained on Kinetics dataset.
}
\end{tabular}}
\end{table}
%-----

%-----
\begin{table}[!t]
\caption{Action recognition performance}
\centering
\resizebox{0.5\textwidth}{!}{%
\label{table:2}
\begin{tabular}{l l c}
\toprule
Model  & Backbone & Weighted F1-score   \\ 
\midrule
\multirow{2}{*}{LSTM} & ESNet & 0.63 \\
                      & RGB I3D  & 0.71 \\
\midrule
\multirow{2}{*}{TCN} & ESNet  & 0.63 \\
                      & RGB I3D  & 0.75 \\
\midrule
\multirow{2}{*}{MS-TCN++} & ESNet  & 0.66 \\
                      & RGB I3D  & 0.78 \\
\midrule
\multirow{2}{*}{MS-TCN} & ESNet  & \textbf{0.71} \\
                      & RGB I3D  & \textbf{0.83} \\                      
%RGB I3D$\dagger$ + LSTM  & 0.71  \\
%RGB I3D$\dagger$ + TCN  & 0.75  \\
%RGB I3D$\dagger$ + MS-TCN++  & 0.78  \\
%\textbf{RGB I3D$\dagger$ + MS-TCN}  & \textbf{0.8}  \\
\bottomrule
%\multicolumn{2}{p{200pt}}
%{ 
%$\dagger$ Pretrained on Kinetics dataset.
%}
\end{tabular}}
\end{table}
%-----

%-----------
\subsection{Action recognition models evaluation}
After identifying the most suitable feature extractors, we aim to test the performance of different action recognition models. Here, we use the best light-weight (ESNet) and conventional (RGB I3D) feature extractor that was identified in Section \ref{sec:backbone_best}. As outlined in Section \ref{subsec:ar}, the action recognition models we consider are LSTM, TCN, MS-TCN and MS-TCN++. A comparison of the models is shown in Table~\ref{table:2}.

From Table~\ref{table:2}, we can see the performance of the TCNs models are better than the LSTM for the ADS action recognition. This result also supports the conclusion of \cite{bai2018empirical}, where TCNs were shown to outperform recurrent neural networks such as LSTM for several tasks. The reason for this performance gain is that TCNs can have an effective longer memory than RNNs. In addition, the I3D+ and ESNet+ MS-TCN model achieved the highest weighted F1-scores of 0.83 and 0.71 on average for conventional and light-weight models, respectively.

It is noted that there is a trade-off between accuracy and computation time. With respect to computation times, ESNet+ MS-TCN model is the best solution because due to its lower complexity (around 2M parameters), while still achieving competitive performance (0.71 weighted F1-score). However, in terms of accuracy, the I3D+ MS-TCN model outperforms other all considered models in this work for  the action recognition task (0.83 Weighted F1-score), yet has higher computational demands.

Although the feature extraction component is a critical component to reach low number of floating-point operations per second (FLOPs), the action recognition model is also challenging to directly adapt to real-world applications due to the temporal modelling structure which requires a sequence of frames to be provided. Thus, real-time performance of these methods is also impacted by how the buffering is performed and the size of the frame buffer used for action recognition. With the proposed backbone and action recognition model, we aim to achieve the best trade-off between accuracy and latency.
We propose the following strategies:
i) Instead of processing images one by one in a video sequence, processing image frames in a batch-wise manner can improve the computational efficiency.
ii) To enable parallel computing in our system, we adopt a first-in-first-out (FIFO) buffer, where the data written into the buffer first comes out of it first. The feature extracted with EsNet are appended to a buffer of size 50, which operates in a sliding window fashion. Content of the buffer is finally sent to the trained MS-TCN model for action identification. Testing on a single Nvidia Geforce RTX 2080Ti and 4 CPU cores, the pipelines takes about 175ms to process a 50 frames of data.
Although our results are promising, additional optimization and model pruning techniques are required for the action recognition model~\cite{gammulle2022continuous}, such as optimizing the computation graph generated in Pytorch with TensorRT, to deploy our pipeline onto an edge-AI system that can be built around, for example a single Jetson AGX Xavier.

\section{Conclusion}
\label{conclusion}
In this paper, we introduce a region-based computer vision system that aims to help clinicians and parents analyze children’s behaviors, and in particular help to identify behaviors associated with Autism Spectrum Disorder (ASD). 

We enhanced an existing dataset to construct a new, expanded dataset of stereotypical children's autism behaviors, using videos recorded in an uncontrolled environment. We investigated several region-based action recognition frameworks for the detection of these characteristic behaviors, including light-weight and standard complexity models. The experimental results show that the proposed framework can recognize autism-related behaviors accurately and robustly. 
Further, from the preliminary experimental results, the light-weight model demonstrated competitive results in comparison to the complex models. We verify the efficacy of the designed strategy for potential real-time processing. In the future, we plan to explore more resource-efficient techniques to speed up the autism diagnosis process, and extend our system to other motor and mental disorders.

\section{Compliance with Ethical Standards}
The experimental procedures involving human subjects described in this paper were approved by the CSIRO Health and Medical Human Research Ethics Committee (CHMHREC).

\bibliographystyle{splncs04}
\bibliography{egbib}

\begin{thebibliography}{10}
\providecommand{\url}[1]{\texttt{#1}}
\providecommand{\urlprefix}{URL }
\providecommand{\doi}[1]{https://doi.org/#1}

\bibitem{ahmedt2019understanding}
Ahmedt-Aristizabal, D., Denman, S., Nguyen, K., Sridharan, S., Dionisio, S.,
  Fookes, C.: Understanding patients’ behavior: Vision-based analysis of
  seizure disorders. IEEE journal of biomedical and health informatics
  \textbf{23}(6),  2583--2591 (2019)

\bibitem{ali2022video}
Ali, A., Negin, F., Bremond, F., Th{\"u}mmler, S.: Video-based behavior
  understanding of children for objective diagnosis of autism. In: VISAPP
  2022-International Conference on Computer Vision Theory and Applications
  (2022)

\bibitem{aihwAutismAustralia}
{Australian Institute of Health and Welfare}: Autism in australia.
  \url{https://www.aihw.gov.au/reports/disability/
  autism-in-australia/contents/about}

\bibitem{bai2018empirical}
Bai, S., Kolter, J.Z., Koltun, V.: An empirical evaluation of generic
  convolutional and recurrent networks for sequence modeling. arXiv preprint
  arXiv:1803.01271  (2018)

\bibitem{de2020computer}
de~Belen, R.A.J., Bednarz, T., Sowmya, A., Del~Favero, D.: Computer vision in
  autism spectrum disorder research: a systematic review of published studies
  from 2009 to 2019. Translational psychiatry  \textbf{10}(1),  1--20 (2020)

\bibitem{carreira2017quo}
Carreira, J., Zisserman, A.: Quo vadis, action recognition? a new model and the
  kinetics dataset. In: proceedings of the IEEE Conference on Computer Vision
  and Pattern Recognition. pp. 6299--6308 (2017)

\bibitem{chen2020action}
Chen, M.H., Li, B., Bao, Y., AlRegib, G., Kira, Z.: Action segmentation with
  joint self-supervised temporal domain adaptation. In: Proceedings of the
  IEEE/CVF Conference on Computer Vision and Pattern Recognition. pp.
  9454--9463 (2020)

\bibitem{national2001educating}
Council, N.R., et~al.: Educating children with autism. National Academies Press
  (2001)

\bibitem{crane2016experiences}
Crane, L., Chester, J.W., Goddard, L., Henry, L.A., Hill, E.: Experiences of
  autism diagnosis: A survey of over 1000 parents in the united kingdom. Autism
   \textbf{20}(2),  153--162 (2016)

\bibitem{cui2021pp}
Cui, C., Gao, T., Wei, S., Du, Y., Guo, R., Dong, S., Lu, B., Zhou, Y., Lv, X.,
  Liu, Q., et~al.: Pp-lcnet: A lightweight cpu convolutional neural network.
  arXiv preprint arXiv:2109.15099  (2021)

\bibitem{deng2009imagenet}
Deng, J., Dong, W., Socher, R., Li, L.J., Li, K., Fei-Fei, L.: Imagenet: A
  large-scale hierarchical image database. In: 2009 IEEE conference on computer
  vision and pattern recognition. pp. 248--255. Ieee (2009)

\bibitem{farha2019ms}
Farha, Y.A., Gall, J.: Ms-tcn: Multi-stage temporal convolutional network for
  action segmentation. In: Proceedings of the IEEE/CVF Conference on Computer
  Vision and Pattern Recognition. pp. 3575--3584 (2019)

\bibitem{gammulle2022continuous}
Gammulle, H., Ahmedt-Aristizabal, D., Denman, S., Tychsen-Smith, L., Petersson,
  L., Fookes, C.: Continuous human action recognition for human-machine
  interaction: A review. arXiv preprint arXiv:2202.13096  (2022)

\bibitem{gammulle2021tmmf}
Gammulle, H., Denman, S., Sridharan, S., Fookes, C.: Tmmf: Temporal multi-modal
  fusion for single-stage continuous gesture recognition. IEEE Transactions on
  Image Processing  \textbf{30},  7689--7701 (2021)

\bibitem{gammulle2021multi}
Gammulle, H., Fernando, T., Sridharan, S., Denman, S., Fookes, C.: Multi-slice
  net: A novel light weight framework for covid-19 diagnosis. In: 2021 IEEE
  International Conference on Autonomous Systems (ICAS). pp.~1--5. IEEE (2021)

\bibitem{gu2018recent}
Gu, J., Wang, Z., Kuen, J., Ma, L., Shahroudy, A., Shuai, B., Liu, T., Wang,
  X., Wang, G., Cai, J., et~al.: Recent advances in convolutional neural
  networks. Pattern Recognition  \textbf{77},  354--377 (2018)

\bibitem{han2020ghostnet}
Han, K., Wang, Y., Tian, Q., Guo, J., Xu, C., Xu, C.: Ghostnet: More features
  from cheap operations. In: Proceedings of the IEEE/CVF conference on computer
  vision and pattern recognition. pp. 1580--1589 (2020)

\bibitem{he2017mask}
He, K., Gkioxari, G., Doll{\'a}r, P., Girshick, R.: Mask r-cnn. In: Proceedings
  of the IEEE international conference on computer vision. pp. 2961--2969
  (2017)

\bibitem{he2016deep}
He, K., Zhang, X., Ren, S., Sun, J.: Deep residual learning for image
  recognition. In: CVPR. pp. 770--778 (2016)

\bibitem{he2019bag}
He, T., Zhang, Z., Zhang, H., Zhang, Z., Xie, J., Li, M.: Bag of tricks for
  image classification with convolutional neural networks. In: Proceedings of
  the IEEE/CVF Conference on Computer Vision and Pattern Recognition. pp.
  558--567 (2019)

\bibitem{hochreiter1997long}
Hochreiter, S., Schmidhuber, J.: Long short-term memory. Neural computation
  \textbf{9}(8),  1735--1780 (1997)

\bibitem{howard2019searching}
Howard, A., Sandler, M., Chu, G., Chen, L.C., Chen, B., Tan, M., Wang, W., Zhu,
  Y., Pang, R., Vasudevan, V., et~al.: Searching for mobilenetv3. In:
  Proceedings of the IEEE/CVF International Conference on Computer Vision. pp.
  1314--1324 (2019)

\bibitem{howard2017mobilenets}
Howard, A.G., Zhu, M., Chen, B., Kalenichenko, D., Wang, W., Weyand, T.,
  Andreetto, M., Adam, H.: Mobilenets: Efficient convolutional neural networks
  for mobile vision applications. arXiv preprint arXiv:1704.04861  (2017)

\bibitem{hu2018squeeze}
Hu, J., Shen, L., Sun, G.: Squeeze-and-excitation networks. In: Proceedings of
  the IEEE conference on computer vision and pattern recognition. pp.
  7132--7141 (2018)

\bibitem{kay2017kinetics}
Kay, W., Carreira, J., Simonyan, K., Zhang, B., Hillier, C., Vijayanarasimhan,
  S., Viola, F., Green, T., Back, T., Natsev, P., et~al.: The kinetics human
  action video dataset. arXiv preprint arXiv:1705.06950  (2017)

\bibitem{kingma2014adam}
Kingma, D.P., Ba, J.: Adam: A method for stochastic optimization. arXiv
  preprint arXiv:1412.6980  (2014)

\bibitem{kojovic2021using}
Kojovic, N., Natraj, S., Mohanty, S.P., Maillart, T., Schaer, M.: Using 2d
  video-based pose estimation for automated prediction of autism spectrum
  disorders in young children. Scientific Reports  \textbf{11}(1),  1--10
  (2021)

\bibitem{krizhevsky2012imagenet}
Krizhevsky, A., Sutskever, I., Hinton, G.E.: Imagenet classification with deep
  convolutional neural networks. Advances in neural information processing
  systems  \textbf{25} (2012)

\bibitem{lea2017temporal}
Lea, C., Flynn, M.D., Vidal, R., Reiter, A., Hager, G.D.: Temporal
  convolutional networks for action segmentation and detection. In: proceedings
  of the IEEE Conference on Computer Vision and Pattern Recognition. pp.
  156--165 (2017)

\bibitem{liang2021autism}
Liang, S., Sabri, A.Q.M., Alnajjar, F., Loo, C.K.: Autism spectrum
  self-stimulatory behaviors classification using explainable temporal
  coherency deep features and svm classifier. IEEE Access  \textbf{9},
  34264--34275 (2021)

\bibitem{lin2014microsoft}
Lin, T.Y., Maire, M., Belongie, S., Hays, J., Perona, P., Ramanan, D.,
  Doll{\'a}r, P., Zitnick, C.L.: Microsoft coco: Common objects in context. In:
  European conference on computer vision. pp. 740--755. Springer (2014)

\bibitem{lord2018autism}
Lord, C., Elsabbagh, M., Baird, G., Veenstra-Vanderweele, J.: Autism spectrum
  disorder. The lancet  \textbf{392}(10146),  508--520 (2018)

\bibitem{ma2018shufflenet}
Ma, N., Zhang, X., Zheng, H.T., Sun, J.: Shufflenet v2: Practical guidelines
  for efficient cnn architecture design. In: Proceedings of the European
  conference on computer vision (ECCV). pp. 116--131 (2018)

\bibitem{negin2021vision}
Negin, F., Ozyer, B., Agahian, S., Kacdioglu, S., Ozyer, G.T.: Vision-assisted
  recognition of stereotype behaviors for early diagnosis of autism spectrum
  disorders. Neurocomputing  \textbf{446},  145--155 (2021)

\bibitem{oord2016wavenet}
Oord, A.v.d., Dieleman, S., Zen, H., Simonyan, K., Vinyals, O., Graves, A.,
  Kalchbrenner, N., Senior, A., Kavukcuoglu, K.: Wavenet: A generative model
  for raw audio. arXiv preprint arXiv:1609.03499  (2016)

\bibitem{pandey2020guided}
Pandey, P., Prathosh, A., Kohli, M., Pritchard, J.: Guided weak supervision for
  action recognition with scarce data to assess skills of children with autism.
  In: Proceedings of the AAAI Conference on Artificial Intelligence. vol.~34,
  pp. 463--470 (2020)

\bibitem{rajagopalan2013self}
Rajagopalan, S., Dhall, A., Goecke, R.: Self-stimulatory behaviours in the wild
  for autism diagnosis. In: Proceedings of the IEEE International Conference on
  Computer Vision Workshops. pp. 755--761 (2013)

\bibitem{rajagopalan2014detecting}
Rajagopalan, S.S., Goecke, R.: Detecting self-stimulatory behaviours for autism
  diagnosis. In: 2014 IEEE International Conference on Image Processing (ICIP).
  pp. 1470--1474. IEEE (2014)

\bibitem{simonyan2014very}
Simonyan, K., Zisserman, A.: Very deep convolutional networks for large-scale
  image recognition. arXiv preprint arXiv:1409.1556  (2014)

\bibitem{szegedy2016rethinking}
Szegedy, C., Vanhoucke, V., Ioffe, S., Shlens, J., Wojna, Z.: Rethinking the
  inception architecture for computer vision. In: Proceedings of the IEEE
  conference on computer vision and pattern recognition. pp. 2818--2826 (2016)

\bibitem{tan2019efficientnet}
Tan, M., Le, Q.: Efficientnet: Rethinking model scaling for convolutional
  neural networks. In: International conference on machine learning. pp.
  6105--6114. PMLR (2019)

\bibitem{wang2019i3d}
Wang, X., Miao, Z., Zhang, R., Hao, S.: I3d-lstm: A new model for human action
  recognition. In: IOP Conference Series: Materials Science and Engineering.
  vol.~569, p. 032035. IOP Publishing (2019)

\bibitem{washington2021activity}
Washington, P., Kline, A., Mutlu, O.C., Leblanc, E., Hou, C., Stockham, N.,
  Paskov, K., Chrisman, B., Wall, D.: Activity recognition with moving cameras
  and few training examples: applications for detection of autism-related
  headbanging. In: Extended Abstracts of the 2021 CHI Conference on Human
  Factors in Computing Systems. pp.~1--7 (2021)

\bibitem{wu2019detectron2}
Wu, Y., Kirillov, A., Massa, F., Lo, W.Y., Girshick, R.: Detectron2.
  \url{https://github.com/facebookresearch/detectron2} (2019)

\bibitem{xie2017aggregated}
Xie, S., Girshick, R., Doll{\'a}r, P., Tu, Z., He, K.: Aggregated residual
  transformations for deep neural networks. In: Proceedings of the IEEE
  conference on computer vision and pattern recognition. pp. 1492--1500 (2017)

\bibitem{yu2021pp}
Yu, G., Chang, Q., Lv, W., Xu, C., Cui, C., Ji, W., Dang, Q., Deng, K., Wang,
  G., Du, Y., et~al.: Pp-picodet: A better real-time object detector on mobile
  devices. arXiv preprint arXiv:2111.00902  (2021)

\bibitem{zhang2018shufflenet}
Zhang, X., Zhou, X., Lin, M., Sun, J.: Shufflenet: An extremely efficient
  convolutional neural network for mobile devices. In: Proceedings of the IEEE
  conference on computer vision and pattern recognition. pp. 6848--6856 (2018)

\bibitem{zunino2018video}
Zunino, A., Morerio, P., Cavallo, A., Ansuini, C., Podda, J., Battaglia, F.,
  Veneselli, E., Becchio, C., Murino, V.: Video gesture analysis for autism
  spectrum disorder detection. In: 2018 24th International Conference on
  Pattern Recognition (ICPR). pp. 3421--3426. IEEE (2018)

\end{thebibliography}
\end{document}